\documentclass[sigconf]{acmart}
\AtBeginDocument{%
  }

\setcopyright{cc}
\copyrightyear{2026}
\acmYear{2026}
\acmDOI{XXXXXXX.XXXXXXX}
\acmConference[FailForward@HRI26]{Failing Forward: 
Design and Deployment Lessons from Real-World Human–Robot Interaction
Workshop @ HRI 2026}{March 16,
  2026}{Edinburgh, Scotland}

\begin{document}

\title{Faulty Coffees: Barriers to Adoption of an In-the-wild Robo-Barista}

\author{Bruce W. Wilson}
\affiliation{%
  \institution{Heriot-Watt University}
  \department{School of Mathematical and Computer Sciences}
  \city{Edinburgh}
  \country{UK}}
\email{bww1@hw.ac.uk}

\author{David A. Robb}
\affiliation{%
  \institution{Heriot-Watt University}
  \department{School of Mathematical and Computer Sciences}
  \city{Edinburgh}
  \country{UK}}
\email{d.robb@hw.ac.uk}

\author{Mei Yii Lim}
\affiliation{%
  \institution{Heriot-Watt University}
  \department{School of Mathematical and Computer Sciences}
  \city{Edinburgh}
  \country{UK}}
\email{m.lim@hw.ac.uk}

\author{Helen Hastie}
\affiliation{%
  \institution{University of Edinburgh}
  \department{School Informatics}
  \city{Edinburgh}
  \country{UK}}
\email{h.hastie@ed.ac.uk}

\author{Matthew Peter Aylett}
\affiliation{%
  \institution{Heriot-Watt University}
  \department{School of Mathematical and Computer Sciences}
  \city{Edinburgh}
  \country{UK}}
\affiliation{%
  \institution{CereProc Ltd.}
  \city{Edinburgh}
  \country{UK}}
\email{m.aylett@hw.ac.uk}

\author{Theodoros Georgiou}
\affiliation{%
  \institution{Heriot-Watt University}
  \department{School of Mathematical and Computer Sciences}
  \city{Edinburgh}
  \country{UK}}
\email{t.georgiou@hw.ac.uk}
\renewcommand{\shortauthors}{Wilson et al.}

\begin{abstract}

We set out to study whether task-based narratives could influence long-term engagement with a service robot. To do so, we deployed a Robo-Barista for five weeks in an over-50's housing complex in Stockton, England. Residents received a free daily coffee by interacting with a Furhat robot assigned to either a narrative or non-narrative dialogue condition. Despite designing for sustained engagement, repeat interaction was low, and we encountered curiosity trials without retention, technical breakdowns, accessibility barriers, and the social dynamics of a housing complex setting. 
Rather than treating these as peripheral issues, we foreground them in this paper. We reflect on the in-the-wild realities of our experiment and offer lessons for conducting longitudinal Human-Robot Interaction research when studies unravel in practice.
\end{abstract}



\received{14 March 2026}

\maketitle

\section{Introduction}

\begin{figure}[t]
  \centering
  \includegraphics[width=0.7\linewidth]{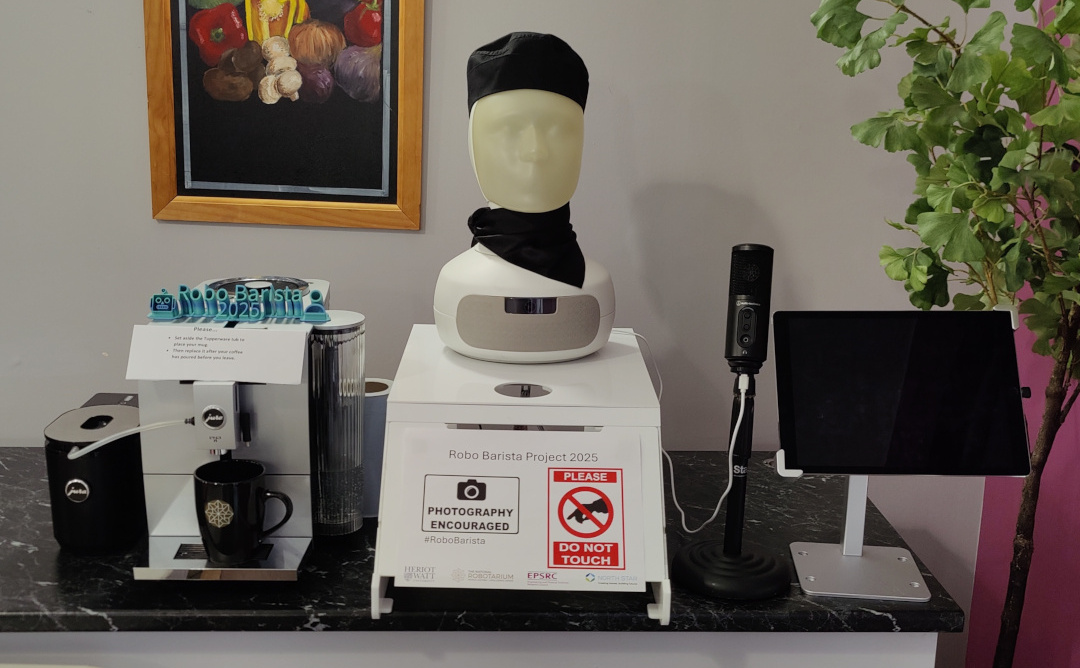}
  \caption{The Robo-Barista Furhat Robot ``Basil'' wearing a bistro uniform, named after the `Faulty Towers' character.}
  \label{fig:graph-user-perceptions}
\end{figure}

Narrative thinking is suggested to be a fundamental way of comprehending our everyday reality \cite{bruner_actual_1986}. With literature surrounding narratives reaching back to Aristotle\cite{lucas_aristotle_1968}, we can define them for the purposes of our task as a series of causal, temporal, and spatial chains, linking together human dialogue fragments \cite{per_persson_supporting_1998}. In human to human communication, narrative communication styles can often produce effects that often do not appear in other more ``expository'' forms of communication \cite{shaffer_usefulness_2018}.
Examples of these effects can be seen across fields, such as in medical research, where narratives can be used to provide examples of others' experiences, acting as a decision aid to provide information, convey empathy, and change behaviour \cite{shaffer_all_2013, bekker_personal_2013}.
However, while prior work looks at implementing other human social characteristics \cite{lim_we_2022} into task-based robots in environments, such as malls \cite{kanda_communication_2010}, bakeries \cite{song_service_2022}, or homes \cite{de_graaf_phased_2018}, little work looks at task-based narratives. Due to this, in human-robot communication, it is not yet clear whether these same effects are present, and whether these effects would be wanted or useful.
Prior work indicates that narratives may influence perceived adaptiveness of robots \cite{wilson_lost_2025}, which describes how humans may think a robot can adapt to their needs over a period of time. Therefore, this follow-up study intended to study this in a natural environment.

In this paper we discuss the deployment of our ``in-the-wild'' Robo-Barista in an over 50's housing complex in Stockton, England. ``Basil,'' named by the residents, resides within the housing complex's bistro, where hot meals and drinks are served to the residents and visitors.  
This location was chosen to attempt to interact with a relatively under-explored user population in a natural environment, and was designed to engage the residents in additional conversation, and free staff time to further engage with the residents in non-serving situations.
The longitudinal study lasted around five weeks, and users were able to receive up to one free coffee per day by interacting with the Furhat robot. Users were assigned one of two conditions (between subjects): narratives or non-narratives. We intended to study the effects of these narratives on participant engagement, attitudes and prejudices towards the robot, and perceptions of the usefulness of the robot. However, while the technical setup of the Robo-Barista was designed to encourage long-term engagement with the system, several unexpected human behavioural usage patterns were uncovered throughout the study, which led to quality and quantity issues. Overall, the experiment was a success for all those involved, and analysis is ongoing with future work, however, this paper shall highlight the aforementioned unexpected usage patterns and lead a discussion around these.



\section{Technical Overview}
The Robo-Barista was set up in the bistro area frequented by residents and staff of the housing complex, and was similar our previously deployed Robo-Barista \cite{lim_feeding_2023} with many technical improvements.
The system consisted of a high-end Bluetooth coffee machine\footnote{https://uk.jura.com/, last accessed: 11/02/2026}, a Furhat robot\footnote{https://furhatrobotics.com, last accessed: 11/02/2026}, an iPad tablet, a 2D barcode scanner, and a laptop. The conversational interaction was implemented using the open-source platform RASA\footnote{https://rasa.com/, last accessed: 11/02/2026} running on the laptop, which handled all conversational processing using the RASA TED Policy \cite{vlasov_dialogue_2019} alongside a set of handwritten rules. Staff at the housing complex asked residents what they would like the robot to be called, and they selected the name Basil, after the `Faulty Towers' hotel manager.

Participant identification is handled by the use of ``loyalty cards'' given to eligible and consenting individuals - a common scheme used in caf\'e settings.
Upon these cards we printed a participant number (a random 4-digit number), and an Aztec code \cite{longacre_two_1997} (a 2D matrix code) representation of the participant number to aid with machine identification. The use of a 2D barcode scanner greatly improved scanning success compared to the system in our previous implementation which used the Furhat robot's onboard camera.

The interaction flow begins with the Robo-Barista detecting an approaching participant. It greets the participant, asking if they have a loyalty card. If they do not yet have a loyalty card, it points them towards recruitment signage. If they do, it prompts them to scan the card on the barcode scanner. The system verifies that the participant number is valid, and ensures that the participant has not already claimed their free coffee that day.
Upon success, if the participant has not previously visited the Robo-Barista, it politely asks them to first complete a short consent form and questionnaire on their prior experience with robots. 
It is at this point that the first conditional exposure is conducted. The robot will present a fact about the local area, in the manner that it is something it has learned recently. This fact is presented in either a narrative or non-narrative format \cite{wilson_lost_2025}, and the user is asked a simple variant of ``did you know this?'' to acknowledge the fact.
After this, the robot continues with the coffee order, offering a ``usual'' if the participant has visited multiple times before, or offering suggestions if asked. 

After coffee selection is confirmed, the robot instructs the participant to place their mug in the coffee machine, and begins brewing the order. The robot asks the participant to answer some quick questions on the tablet about their interaction that day while it is brewing the drink. This timing is convenient as while the coffee machine is brewing it is too loud for the speech recognition of the robot to accurately hear the participant. 

Next, a second conditional exposure is performed. Once participants have completed the questionnaire and received their drink, the robot is preloaded with knowledge of the activities taking place in the housing complex on that day of the week (for example a quiz or bingo). If an activity is happening in the morning or afternoon, it will remind participants, and ask if they are planning on attending. Otherwise, it will let participants know that there are no further activities planned for today, and asks if they will be attending an activity tomorrow. These reminders are given in either a narrative or a non-narrative format.
Finally, the robot thanks participants for their time, and waits for a new participant to approach.

As the deployment of the system was far from the researchers location, remote capabilities were included in the design. The robot and laptop could both be remotely fixed when errors occurred using Cloudflare zero-trust tunnels\footnote{\url{https://developers.cloudflare.com/cloudflare-one/networks/connectors/cloudflare-tunnel/} last accessed: 11/02/2026}. Additionally, the system automatically alerted the research team both during start up and throughout the day if any error occurred, who could then liaise with the bistro team at the housing complex to attempt any fixes.



\section{Discussion and Lessons Learned}

Throughout the five weeks that the Robo-Barista was deployed, we gathered continuous data from the deployment setup itself, including direct participant feedback mid-interaction, but also from the staff at the deployment location who relayed any practical issues back to us.
This was then combined with exit questionnaires to produce our dataset. 
In this section, we will discuss challenges we encountered throughout the deployment period, leading to our dataset being incomplete to answer the initial hypotheses. 
Therefore, this section does not intend to be a formal results section, nor does it include a complete thematic analysis, but a discussion of first-hand experiences from participants, staff, and the researchers, in the form of feedback and lessons learned. Future work will include a redeployment of an improved system to gather a second dataset, and a formal analysis of this studies dataset to contrast.

Forty-four coffees were served by the Robo-Barista to 32 participants (excluding demonstration coffees for guests) throughout its five weeks of operation. While we wanted to gather longitudinal feedback over time, only one participant received four coffees, two participants received three coffees, and five participants received two coffees. The remaining 23 participants only used the system a single time. During interviews with the bistro staff, it was noted that there were many instances of lost loyalty cards and people sharing loyalty cards, which further jeopardises the quality of the quantitative data collected by the robot.
This was then amplified when we collected exit interviews and only received seven exit interviews (five participants, two staff)
Although qualitative data may not be easily generalised from, we did gather some glimpses into our participants' views, which we will discuss below.


\subsection{High Attrition Rate}



Participant 1 said: ``I don't drink a lot of coffee - I need to fancy one,'' and followed up with ``I'm not a big coffee drinker anyway.'' Participant 2 had a similar sentiment: ``I think I would use it, just the odd time, just not regularly - if I was really fancying [a coffee].'' 
The bistro staff further reflected this when asked: ``Yeah I had one a couple of times, but... then you just think, I'll just [make] my own, in terms of quickness to get it'', which participant 4 agreed with, commenting that ``I would continue to use it in a more appropriate environment, and if it was a bit quicker to use and produce a coffee.''

Staff talked further on the topic of resident interest: ``At first, [the residents] wanted to try it, obviously we needed to help them, but then after they tried it, they didn't want to try again... A few of them did, but they came and got a coffee, then walked away.'' Asking what gave them this impression, staff mentioned that ``if one person got it, the others didn't say `could I have one', they just walked away,'' continuing with ''there wasn't a whole lot of conversation about it. They asked what it was for, we told them it was for [Researcher's doctorate], and they were like oh okay, well who's [Researcher]?'' They noted that it was a ``shiny new toy in the corner,'' but after the first interaction, ''near the end, I think they'd seen it, done it, got the t-shirt. A lot of them started the interactions, then just left.'' The staff however did finish by saying ``if you mentioned Basil, they know who you were talking about at least,'' continuing that ``I think that it made a difference that it was free, they had a go. If it cost a pound, they probably wouldn't have.''

Overall, these observations give the impression that participants saw the Robo-Barista as something ``have a go at,'' rather than use regularly, and a single curious interaction did not spark the desire to engage in future interactions. Many factors seem to be influencing this return rate, such as quickness of the dispensing - where participants may be used to a much faster service from the bistro staff, versus the longer interaction typical in a coffee shop, or a misunderstanding of the use for this system - again due to familiarity.
The lack of visible social momentum and limited conversational engagement further suggest the system did not integrate into existing social routines. 
Together, this highlights the importance of perceived usefulness and social embedding when introducing autonomous service agents into community environments.
Matters like these can be resolved with stronger interaction design, however, this goes hand in hand with the reliability of the system design.

\subsection{Usability for Me, not for Thee}

After the previous deployment of the Robo-Barista, many technical changes were made to the system in an attempt to increase ease of use and rate of data collected. These included: switching to a dedicated loyalty card scanner rather than using the Furhat's onboard camera to decrease the likelihood of the Furhat losing tracking on the participant; presenting all questionnaires on the integrated tablet system while the coffee was brewing to increase rate of questionnaire answering rather than users walking away the second their drink was made; allowing for almost full remote repair to the interaction stack to fix any faults. This did solve the issue from the previous system where some users would answer no questionnaires at all, including the mandatory sign-up questionnaires displayed via QR codes - however, when attrition was so high and so few participants interacted more than once, it is unclear whether these changes made a meaningful difference to ease of use or data collection rate. Certainly, it could have made a negative impact on both, and must be investigated in future work.

We asked participants and staff what improvements could be made to the system. Participants 2 and 3, who were interviewed at the same time as they were sitting together, immediately started talking about the uptime of the system, with participant 2 noting that ``they couldn't seem to get it going at all, we'd do so much and then it'd go off. There wasn't a continuation with it at all... and we'd have to wait and try and get a coffee.'' Participant 3 continued: ``I tried, it never came out at all, I never got one - poor fella sat there, and I got nothing,'' commenting on their one and only interaction with the system where it could not hear them. Participant 5 had a similar no-coffee experience, noting that ``I took [my] cup away too early! So I [didn't get anything].'' Participant 4 mentioned that ``the coffee was only lukewarm at first, so I suggest it is served a lot hotter,'' which the bistro staff also alerted us to early on in the project. We attempted to fix this, however it must have remained a prominent issue for this participant as they only received two coffees. Finally, participant 1 remarked that ``I needed help a few times with it, from the bistro staff - to get the coffee in it.''  

Being unable to hear was a common topic, both for the robot and the participants. The bistro staff noted that ``during lunchtime you could hardly hear him and I think he struggled a bit,'' continuing that ``maybe if he was just a bit louder, as [the staff] often had to repeat. A lot of times Basil did repeat himself, but... [the participants] would just be like `oh it doesn't matter,' and [leave].''
Another issue brought up by the bistro staff revolved around the Robo-Barista Menu.
The Jura ENA 8 coffee machine is capable of producing espresso, cappuccino, macchiato, and other barista style drinks. However, as a member of staff at the bistro notes, ``most of [the residents] around here know two types of coffee: black and white.'' 

Several comments were made about the tablet device we used to collect feedback. The bistro staff summarised these well, stating that ``the text on the screen was too small, and a lot of [the residents] haven't got [any] idea about those forms, on the iPad, the touch, or anything like that. Maybe if Basil could have asked the questions?'' The staff continued later: ``we have some people with [sight difficulties]... it's very time consuming [to help them] with that.'' While we intended to make it easier to collect our data, we inadvertently made it more difficult for this group to interact with the system, and increased the workload of the bistro staff who graciously stepped in to help participants interact with the screen. 

These observations indicate the need for further pilot testing in our target environment and with our target group.
While technical modifications improved functionality and questionnaire completion, user-facing breakdowns including failed dispensing, acoustic challenges, and tablet accessibility barriers - likely undermined trust formation and therefore lowered the likelihood of returning participants. Moreover, differences in mental models between expected menu options versus reality may play a large role in perceived usefulness of the system, and fewer options may greatly improve adoption.
Service systems are judged on their functional consistency, and a single-point failure in a first interaction could permanently reduce willingness to retry. In our previous study, we saw that some participants would adapt to the system by, for example, speaking louder, or standing differently. However, if, as we previously discussed, participants saw the system only as a curiosity, then a single failure is destined to see them never return. 

Additionally, the system inadvertently increased staff workload, highlighting the importance of considering invisible labour in robot deployment. If a system increases staff effort, rather than reducing it or alleviating stress, long-term institutional support may decline.
With a combination of further research on targets, and pilot testing in-situ, these issues should be ironed out long before participants first impressions are made. While the interaction \textit{dialogue} was crafted for our target group, the interaction \textit{design} requires further tailoring to be more suitable.






%



The system also had a sustained period of downtime during the second week of deployment when the coffee machine broke down and had to be repaired by a member of the research team. Staff were asked to comment on this, but first remarked on their initial experience when the instructions on turning on the system in the morning were missing a step: ``the first time when it broke, obviously, it was just new, and we didn't know [what to do]. But once we knew, just flick the button at the back, it'll reset it, job done.'' Prompting how this made them feel about the full breakdown later, a staff member who had not turned the system on before said ``obviously it was like my first day on it, setting it up, and it's like, well what the f*ck do I do.''

These breakdowns highlight the operational fragility of auto-nomous systems deployed in real-world settings, especially where downtime occurs when momentum has not yet formed.
Beyond technical repair, the incidents exposed gaps in procedural documentation, and uneven distribution of troubleshooting knowledge. 
Confidence of staff operating in-the-wild studies is critical to deployment success, and uncertainty during faults may indirectly shape participant perceptions of the overall systems reliability.
Future deployments should treat monitoring, support availability, and detailed independently-verified instructions, as integral components of system design rather than secondary considerations to minimise the consequences of malfunctions.

\vspace{-5pt}
\subsection{Humans being Humans}

Further questioning in the direction of how the interaction compared to a human barista gleamed some details. Participant 2 said ``It didn't really appeal to me at all, I'd rather have a human, to speak to - the contact, that's where it was missing the human.'' Participants 3, 1, and 5 did not comment on the difference, instead agreeing that the Robo-Barista was ``a nice thing to try and do,'' but did not specify whether they preferred a human or not. 
Participant 1 then continued down a different track ``I don't really come down here too much, to the bistro - the only time we come down here is for lunch - and the [bistro staff] here are very busy if we needed help - it was difficult timing wise.'' Participant 3 agreed, and noted that by dinner time the system was off anyway.
They then continued, and interestingly participant 1 spoke about another resident who didn't take part in the experiment. ``[Resident name] is always there all the time, I try to avoid them [laughs].'' Participant 3 agreed, ``he's all over the place, [what he says] might be fact or it might be fiction.''

The reality of in-the-wild systems like our Robo-Barista is that we cannot control humans and where they linger. It also shines a spotlight in the darkness, where many interactions between humans could be occurring, influencing usage, but that we know nothing about. Tracking more data around the robot could help gain further insights, however, incredible care must be taken surrounding the ethical impact of users who are simply nearby the system.
As most participants did not comment on a difference in human-likeness, are we preaching to the choir, where the system becoming more human is turning our single interactees? Or does this indifference indicate a lack of emotional investment? Future deployments must aim to gather exit feedback from as many users as possible, potentially even those who never completed a full interaction, but expressed interest then decided against it later. 

The bistro staff had differing opinions on the comparison to a human barista, with one noting: ``I think human baristas are just a case of they want you served in, out, and gone - In quieter times if you could hear [Basil], you could hear him chatting more.''
The bistro staff later continued to talk about the nature of the chit-chat: ``It [is nice] to the likes of the residents, this is their only interaction when they're coming in for their lunch. It's not a case of them staying in, they have nobody to visit them, so the only case they have company is when they're coming in here.''

Robotics has strong potential in the care and ageing sector, however, great care is required when a robot has the potential to ``replace'' a human interaction, especially when the bistro functions as a key site of social contact for residents. Autonomous systems in such contexts must be designed to supplement, rather than displace, human interaction. While our system intended to measure narrative effects and free up staff time to have more meaningful conversations with residents, in the end it used more staff time - which inadvertently gave bistro staff even more time with residents. In a confusing mess of meaningful interaction time and the worries of replacing human contact, future work should more deliberately capture these nuances of autonomous system service situations.

\section{Conclusions}

This paper explored how our Robo-Barista was received within a real-world community setting, drawing on the experiences of residents and staff. While initial curiosity was evident, sustained engagement was undermined by reliability, human social and environmental fit, and accessibility. Early breakdowns, contextual mismatches, and the existing human ecology of the housing complex all influenced adoption, and compared to our previous work, technical refinements alone were insufficient to secure continued use. 
These findings do not seek to generalise beyond this setting, but instead offer situated insights into the practical and relational complexities of deploying in-the-wild robot systems.






\begin{acks}
The Robo-Barista in the Community Hub project was funded by NorthStar Housing Group, and facilitated by The National Robotarium.
A huge thank you goes out to the bistro staff at the deployment location for continuously maintaining the setup and helping participants with any questions they had.
Bruce W. Wilson's PhD funding is provided by Engineering and Physical Sciences Research Council (EPSRC) Centre for Doctoral Training in Robotics and Autonomous Systems (EP/S023208/1). 
\end{acks}

\bibliographystyle{ACM-Reference-Format}
\balance
\bibliography{references}

\end{document}